\documentclass{ieeeaccess}
\usepackage{cite}
\usepackage{amsmath,amssymb,amsfonts}
\usepackage{graphicx}
\usepackage{algpseudocode}
\usepackage{textcomp}
\usepackage{algorithm}
\usepackage{booktabs}
\usepackage{bm}
\usepackage{stfloats}
\usepackage{multirow}
\usepackage{multicol}
\usepackage{makecell}
\usepackage{xspace}
\usepackage{mathtools}
\usepackage{latexsym,mathtext} 
\usepackage{adjustbox}
\usepackage{url}

\makeatletter
\AtBeginDocument{\DeclareMathVersion{bold}
\SetSymbolFont{operators}{bold}{T1}{times}{b}{n}
\SetSymbolFont{NewLetters}{bold}{T1}{times}{b}{it}
\SetMathAlphabet{\mathrm}{bold}{T1}{times}{b}{n}
\SetMathAlphabet{\mathit}{bold}{T1}{times}{b}{it}
\SetMathAlphabet{\mathbf}{bold}{T1}{times}{b}{n}
\SetMathAlphabet{\mathtt}{bold}{OT1}{pcr}{b}{n}
\SetSymbolFont{symbols}{bold}{OMS}{cmsy}{b}{n}
\renewcommand\boldmath{\@nomath\boldmath\mathversion{bold}}}
\makeatother

\def\BibTeX{{\rm B\kern-.05em{\sc i\kern-.025em b}\kern-.08em
    T\kern-.1667em\lower.7ex\hbox{E}\kern-.125emX}}

\begin{document}
\history{Received 20 July 2024, accepted 1 August 2024, date of publication 7 August 2024, date of current version 19 August 2024.}
\doi{10.1109/ACCESS.2024.3440006}

\title{Table-to-Text Generation with Pretrained Diffusion Models}
\author{\uppercase{Aleksei S. Krylov}\authorrefmark{1,2}, 
\uppercase{Oleg D. Somov}\authorrefmark{1,3}}

\address[1]{Moscow Institute of Physics and Technology, Moscow, Russian Federation}
\address[2]{Sber, 117312 Moscow, Russia}
\address[3]{Artificial Intelligence Research Institute (AIRI), Moscow, Russian Federation}

\tfootnote{``This work was supported in part by the Artificial Intelligence Research Institute.''}

\markboth
{Krylov \headeretal: Preparation of Papers for IEEE TRANSACTIONS and JOURNALS}
{Krylov \headeretal: Preparation of Papers for IEEE TRANSACTIONS and JOURNALS}

\corresp{Corresponding author: Aleksei S. Krylov (e-mail: krylov.as@phystech.edu).}

\begin{abstract}

Diffusion models have demonstrated significant potential in achieving state-of-the-art performance across various text generation tasks. In this systematic study, we investigate their application to the table-to-text problem by adapting the diffusion model to the task and conducting an in-depth analysis. Our experiments cover multiple aspects of diffusion models training. We explore sampling strategy influence by inducing recent diffusion model accelerator DPM-Solver++ into our core model.  We have tested different prediction aggregation methods, like ROVER and Minimum Bayes-Risk (MBR).  Our studies cover the impact of the pre-training phase in diffusion models and the generation length constraints influence. We also have compared diffusion model generation with auto-regressive text-to-text models with different temperature settings for diversity evaluation. Our key observation is that diffusion models demonstrate the balance between quality and diversity while auto-regressive text-to-text models are not successful at handling both at the same time. 
Furthermore, we found out that to achieve the highest quality possible, it is preferable to use a regular sampler with the strictest length constraint to create multiple samples, and then use MBR to aggregate the predictions. However, if you are prepared to give up high level of diversity and to accelerate the process, you can also utilize a fast sampler DPM-Solver++. Our findings reveal that diffusion models achieve comparable results in the table-to-text domain, highlighting their viability in the table-to-text challenge as a promising research direction. 

\end{abstract}

\begin{keywords}
Machine learning, diffusion models, deep learning, data-to-text, table-to-text, natural language processing, artificial intelligence.
\end{keywords}

\titlepgskip=-21pt

\maketitle

\section{Introduction}
\label{sec:introduction}

\PARstart{D}{iffusion} models, initially prominent in the vision \cite{ho2022imagen}, have shown remarkable potential in text generation tasks. These iterative generative models are trained to recover corrupted data through a multi-step denoising process, refining samples from pure noise to produce high-fidelity and diverse outputs. Inspired by their success in vision \cite{ho2022imagen} and audio  \cite{kong2021diffwave}, researchers are exploring diffusion models for text generation \cite{li2022diffusionlm, gong2023diffuseq, lin2023text}, where they introduce a novel noising paradigm and a distinct training objective. This approach offers an alternative to traditional token prediction methods, promising enhanced language modeling capabilities and paving the way for advancements in generating coherent and contextually appropriate text.

Data-to-text generation is the task of generating a target textual description conditioned on source content in the form of structured data. Table-to-text generation involves creating textual descriptions from tables. Examples include generating sentences from historical data, attractions descriptions, football game summaries from box score statistics, and fun facts from Wikipedia info tables \cite{parikh2020totto}.

Current experiments demonstrate that state-of-the-art neural models struggle to generate satisfactory results, despite the high quality of the training data. Therefore it is necessary to explore other generation methods.

In turn, we propose a diffusion model for this task, because it is known for generating highly diverse outputs, robust generative process, and advanced prediction aggregation techniques, and in natural language processing, in such tasks as summarization, paraphrasing, open domain dialogue, diffusion models showed increased text diversity, while maintaining a comparable accuracy level to modern text-to-text solutions \cite{gong2023diffuseq, lin2023text}.

In our work, we have applied the diffusion model GENIE \cite{lin2023text}  with a fast sampler \cite{gong2023diffuseqv2} to the table-to-text ToTTo challenge \cite{parikh2020totto}, because there isn't a pre-trained diffusion architecture with weights other than this one, and pre-training of own architecture is excessively costly and resource-intensive. For the comparison with our solution, we have used models of relevant sizes from the T5 family \cite{raffel2023exploring, kale-rastogi-2020-text}, which are auto-regressive text-to-text models. 

As a core part of the paper, we have carried out extensive research on the main parts of diffusion models.
We did an in-depth impact analysis of the pre-training phase and generation length of the diffusion model. We replaced the original sampler with efficient DPM-Solver++ \cite{gong2023diffuseqv2} and evaluated its performance on our task. We compared the diffusion model diversity with different temperature sampling of the T5 model \cite{raffel2023exploring, kale-rastogi-2020-text}.

Our results show that diffusion models can achieve similar scores in terms of accuracy and diversity in comparison to auto-regressive text-to-text models. 

The contribution of this paper is the following:

\begin{itemize}

\item We propose an approach for the table-to-text problem that leverages an pretrained diffusion model.

\item We compare the proposed method to the auto-regressive baselines and experimentally show that it is effective both in terms of the quality and diversity. 

\item  We adapt fast DPM-Solver++ for GENIE model and compare with regular architecture to test impact to the quality and diversity.

\item We discover that trained from scratch diffusion models outperform autoregressive models on the table-to-text issue. 

\item We assess several features of diffusion models, including the influence of output length limits, different aggregation techniques, and also the impact of sampling temperature to auto-regressive baselines.

\end{itemize}

\section{Related Work}

\subsection{Data-to-text challenge}
Data-to-text is a task of describing structured data adequately and fluently. The three most common datasets for this domain are WebNLG \cite{gardent-etal-2017-webnlg}, MultiWoz \cite{budzianowski2020multiwoz}, and  ToTTo \cite{parikh2020totto}. WebNLG \cite{gardent-etal-2017-webnlg} is a dataset with approximately 18K graphs of subject-object-predicate triplets and their textual descriptions. MultiWoz \cite{budzianowski2020multiwoz} consists of 10K human-human dialogues for developing task-oriented dialogue systems with approximately 56K training samples. The ToTTo dataset \cite{parikh2020totto} is a corpus of over 120K Wikipedia tables paired with natural language descriptions. We conducted our experiments on the ToTTo dataset\cite{parikh2020totto}, because it is the biggest and most modern dataset for the data-to-text task. \\
Recent research utilized mostly neural solutions for end-to-end generation from a linearized table to its summary. After the triumph of extensively pre-trained sequence-to-sequence Transformer models, modern state-of-the-art systems utilize these models for table-to-text generation, in which the input table is converted into a textual sequence through linearization. Examples of such models include the pre-trained language model Bert-to-Bert \cite{Rothe_2020} and T5 models \cite{kale-rastogi-2020-text}. Condensing a table's extensive information into one sentence is challenging, but focusing on key details enables selective descriptions, so \cite{wang2022robust} suggest T5 with structural attention and special position encoding. This helped improve the generation results, but not significantly. Also, some researchers proposed plan-based methods, such as PlanGen \cite{su2021planthengenerate} and Content Planner \cite{puduppully2019datatotext} to control the structure of the generated output. Eventually, an CoNT \cite{an2023cont} developed a new Contrastive Neural Text generation framework and validated it on data-to-text tasks, achieving state-of-the-art quality among base-size models. However, none of existing methods achieved ideal quality, so this still remains an unsolved problem. In addition, there are no works on table-to-text diffusion models and no in-depth exploration of diversity metrics in that challenge, and we decided to do it.

\subsection{Diffusion models in NLP}
Diffusion models have demonstrated significant potential in achieving state-of-the-art performance across various text generation tasks \cite{kong2021diffwave, ho2022imagen}.
At first text-to-text diffusion models were discrete, for example, each token can, with some probability, be noisy or replaced with a random token \cite{hoogeboom2022autoregressive, hoogeboom2021argmax, austin2023structured}. 
Then Diffusion-LM \cite{li2022diffusionlm} appeared as the diffusion model in continuous space for unconditional generation, which maps discrete tokens into continuous latent variables, achieving more complex controllable text generation through continuous diffusion. DiffuSeq \cite{gong2023diffuseq} proposed a new method for conditional text generation. This represents a BERT-like encoder where it is fed concatenated input and additionally noisy output. 

In GENIE \cite{lin2023text} was presented a pre-trained encoder-decoder diffusion transformer which improved text generation quality. 

Lately, the most notable advancements in diffusion text generation models are - Difformer \cite{gao2023difformer}  with an anchor loss to regularize the embeddings and stabilize the training simultaneously, and Diffuseq-v2 \cite{gong2023diffuseqv2} which proposed a decision for increasing the speed of training in 4 times and speed of inference in 800 times.

\section{Method}

\Figure[t!](topskip=0pt, botskip=0pt, midskip=0pt){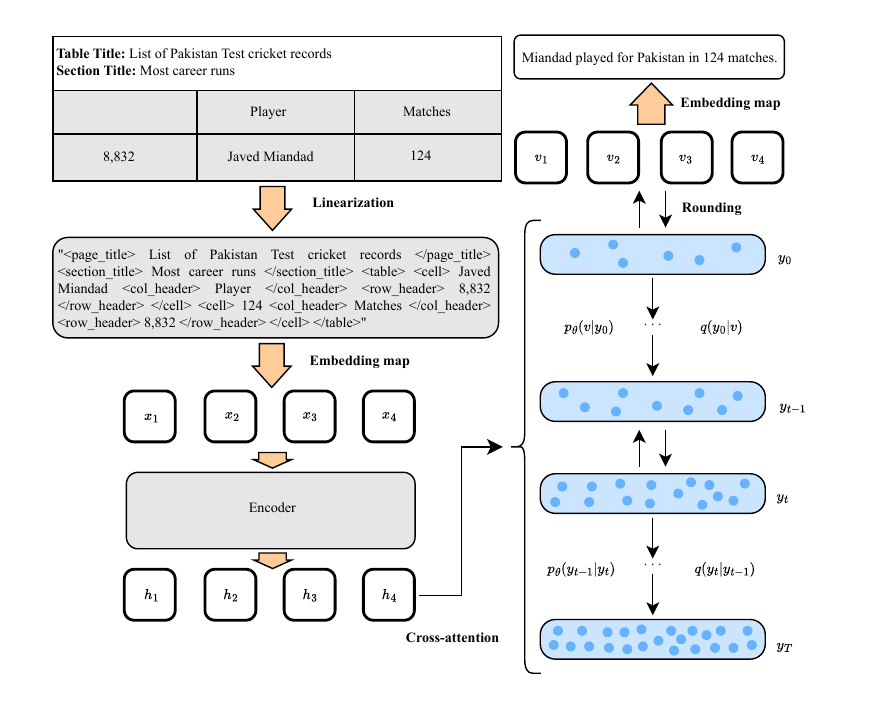}
{ \textbf{We take the linearized source table as the input of the Encoder to obtain the hidden information embeddings which interact with the Language Diffusion Model through cross attention. The Language Diffusion Model deletes the random Gaussian noise from the output text embeddings through the iterative denoising and grounding process. Finally, we map received embeddings to tokens and concatenate them to get the output summary of the input table.}\label{fig:mesh1}}

The proposed method is shown in the Fig. \ref{fig:mesh1} and is described in detail below.

\subsection{Data preparation}

At first, we translate tables into sequences. We use the standard linearization method, proposed by \cite{parikh2020totto}. Then we represent a transformed table as vector $\textbf{x} = \{x_1, x_2, x_3, ..., x_n\}$ with \textbf{n} tokens $x_i$, where $i$ is the token's position. The target sequence is also represented as vector $\textbf{y} = \{v_1, v_2, v_3, ..., v_m\}$ with \textbf{m} tokens $v_i$, where $i$ is the token's position.

\subsection{Diffusion model}

Diffusion models are a type of iter-NAR model \cite{gong2023diffuseq} where a series of intermediate sequences $\mathbf{v}^y$ are introduced over $T$ iterations:

\begin{multline}
\label{eq:iter-nar}
p_\text{iter-NAR}(\mathbf{v}^y_{1:n}|\mathbf{w}^x) =\sum_{\mathbf{v}^y_1,\ldots,\mathbf{v}^y_{T-1}}\Big{[}{\underbrace{\prod_{i=1\ldots n}{p(\mathbf{v}^y_{1,i}|\mathbf{w}^x)}\vphantom{\prod_{t=1..T-1}}}_{\text{initial prediction}}}\cdot \\
\cdot
{\underbrace{\prod_{t=1..T-1}{\prod_{i=1\ldots n}{p(\mathbf{v}^y_{t+1,i}|\mathbf{v}^y_{t,1:n},\mathbf{w}^x)}}}_{\text{progressive full-context prediction}}}\Big{]}.
\end{multline}

We use the pre-trained encoder-decoder diffusion model GENIE \cite{lin2023text} for conditional generation $p(y|x)$. This model contains forward and reverse discrete-time Markov processes. 

\subsubsection{Forward process}
In the forward process, we gradually perturb embeddings of input tokens according to a variance scheduler $\beta_0, \beta_1, ..., \beta_{T}$, during $T$ iterations, while we get standard Gaussian noise. The diffusion process starts with the initial state $x_0$ at time step $t = 0$, where $x_0$ is sampled from the Gaussian distribution of the original data $q(y_0|v) = \mathcal{N}(Emb(v), \beta_0 I)$. At the time step $t + 1$, the latent variable $y_{t+1}$ is only determined by the $x_t$ at time $t$, expressed as:
\begin{equation}
q(y_{t+1} | y_{t}) = \mathcal{N}(y_{t+1}; \sqrt{1 - \beta_{t+1}}y_t, \beta_{t+1} I)
\end{equation}
\subsubsection{Reverse process.}
The reverse diffusion process is needed to recover the original $y_0$ by denoising $y_t$:
\begin{equation}
p(y_{t} | y_{t+1}) = \mathcal{N}(y_{t}; \mu_{\theta}^{t}, \sigma_t)
\end{equation}
\begin{equation}
\mu_{\theta}^{t}(y_t, y_{\theta}) = \frac{\sqrt{\alpha_{t}}*(1 - \bar\alpha_{t-1})}{(1 - \bar\alpha_{t})}*y_{t} + \frac{\beta_{t}*\sqrt{\bar\alpha_{t-1}}}{1 - \bar\alpha_{t}}*y_{\theta}
\end{equation}
\begin{equation}
\sigma_t^2 = \frac{1 - \bar\alpha_{t}}{1 - \bar\alpha_{t+1}}*\beta_{t+1}
\end{equation}
where $\alpha_{t+1} = 1 - \beta_{t+1}$, $\bar\alpha_{t+1} = \prod_{i=1}^t\alpha_{i}$ and the vector $y_{\theta}$ is predicted by a neural network parameterized by $\theta$.

\subsubsection{Training step}
We want to minimize the distance between forward and reverse trajectories to receive a transformer-type model that is capable of removing noise and predicting $p(y|x)$ in the iter-NAR strategy.
We use the variational lower bound to optimize the negative log-likelihood $\mathbf{E}[-logp_{\theta}(y_0)] <= \mathcal{L}_{VLB}$. The final objective is a combination of several KL-divergence and entropy terms following \cite{gong2023diffuseq}:
\begin{multline}
\mathcal{L}_{VLB} = \mathcal{L}_T + \mathcal{L}_{T-1} + ... +  \mathcal{L}_0 = \\
\mathbf{E}_{q(y_{1:T}|y_0, x)}[log\frac{q(y_T|y_0, x)}{p_{\theta}(y_T)} + \sum_{i=2}^{T}log \frac{q(y_{t-1}|y_0, y_t, x)}{p_{\theta}(y_{t-1}|y_t, x)} \\+ log \frac{q(y_0|v)}{p_{\theta}(y_0|y_1)} - log p_{\theta}(v|y_0)] 
\end{multline}
\begin{multline}
\mathcal{L}_t = \mathbf{E}_{q(y_{1:T}|y_0, x}[Const*||\mu_t(y_t, y_0) - \mu_{\theta}(y_t, t)||^2] = \\
Const*\mathbf{E}_{q(y_{1:T}|y_0, x}[y_0 - f_{\theta}(y_t, t)||^2]
\end{multline}

We sample $t$ from the distribution $r_t \sim \sqrt{\mathbf{E}[\mathcal{L}_t^2]}$, $\sum_{i=0}^{T-1} p_t = 1$, thus we can sample more often those steps where we have the biggest loss.

\subsubsection{Inference step}
First of all, we score our input table with the help of an encoder. After that, we generate noise from the Gaussian distribution and consequentially transform it into the target description, as described in Algorithm \ref{algo:main_algorithm}.

\begin{algorithm}[t!]
\caption{Inference algorithm}
\label{algo:main_algorithm}
\begin{algorithmic}[1]
\State \textbf{Input:} Linearized table as a vector of tokens.
\State \textbf{Output:} Summary of table as text.

\State Transform input tokens to embeddings.
\State Pass them through the encoder and get hidden states.
\State Initialize input for diffusion decoder $\mu = 0 \text{ and } \sigma = I$.

\For{$t=T-1...1$}
    \State $y_{t} = \text{ Sample from } \mathcal{N}(\mu_{\theta}^{t}(y_{t+1}, y_0), \sigma_{t+1})$
    \State We pass $y_t$ to the decoder and with the help of hidden states from the encoder non-autoregressive generate $z_{\theta}$
    \State $y_{\theta}(y_{t}, t) = \text{Quantize } z_{\theta}(y_{t}, t) \text{ to nearest embeddings}$
    \State $\mu_{\theta}^{t-1}(y_t, y_0) = \frac{\sqrt{\alpha_{t}}*(1 - \bar\alpha_{t-1})}{(1 - \bar\alpha_{t})}*y_{t} + \frac{\beta_{t}*\sqrt{\bar\alpha_{t-1}}}{1 - \bar\alpha_{t}}*y_{\theta}(y_{t}, t)$
    \State $\sigma_{t-1}^2 = \frac{1 - \bar\alpha_{t-1}}{1 - \bar\alpha_{t}}*\beta_{t}$
\EndFor
\end{algorithmic}
\end{algorithm}

\section{Experiments}

\begin{table*}

\centering

\begin{tabular}{l|l}
\multicolumn{2}{l}{\textit{\textbf{Input Table: }}}\\
\toprule
\multicolumn{2}{l}{\textbf{Table Title: } A. J. Hawk }\\
\multicolumn{2}{l}{\textbf{Section Title: } Career statistics }\\
\midrule
  {row index}& \textbf{TOTAL} \\
  \midrule
 {            } &  {119} \\ 
\bottomrule
\end{tabular}
\newline
\vspace*{0.1 cm}
\newline

\centering
\begin{tabular}{l|l}
\toprule
\multicolumn{2}{l}{\textit{\textbf{Original sentence}: In his rookie season, Hawk led with 119 total tackles.}}\\
\midrule

\textbf{T5-small}  & \textbf{Ours} \\
 {hawk had 119 tackles.} &  {a. j. hawk played 119 games.} \\ 
 {hawk had 119 total tackles} &  {hawk led the team with 119 runs.}\\ 
 {hawk had 119 career total tackles.} &  {hawk scored 119 home runs.} \\
 {hawk scored 119 runs.} &  {a. j. hawk had 119 hits.}\\
{hawk scored 119 strikeouts.} &  \makecell[l]{hawk earned a total of 119 runs batted in during\\ his career.}\\
\bottomrule
\end{tabular}
\caption{Sample outputs in ToTTo \cite{parikh2020totto} development set, conditioned on the same $\textbf{x}$.}
\label{tb:case}
\end{table*}

\begin{table*}
\centering
\begin{tabular}{lllllll} \hline
\multirow{2}{*}{Model} & \multicolumn{2}{c}{Overall} & \multicolumn{2}{c}{Overlap}  & \multicolumn{2}{c}{Non-Overlap} \\ 
                      & BLEU$\uparrow$        & PARENT$\uparrow$        & BLEU$\uparrow$        & PARENT$\uparrow$        & BLEU$\uparrow$          & PARENT$\uparrow$          \\ \hline
BERT-to-BERT \cite{Rothe_2020} & 44.0 & 52.6 & 52.7 & 58.4 & 34.8 & 46.7   \\
T5-Small \cite{kale-rastogi-2020-text} & 45.7 & 55.9 & 53.9 & 60.4 & 37.7 & 51.6   \\
T5-Base \cite{kale-rastogi-2020-text} & 47.7 & 57.1         & 56.1        & 61.8         & 39.6          & 52.6           \\
T5-Large \cite{kale-rastogi-2020-text}                   & 48.1        & 57.3         & 56.8        & 62.0         & 39.8          & 52.8           \\
T5-3B \cite{kale-rastogi-2020-text}                   & 48.4        & 57.8          & 56.7        & 62.4          & 40.4          & 53.3      \\ 
T5-base-CONT \cite{an2023cont}                   & 49.2        & 59.4         & -        & -         & 41.5          & 55.0           \\
T5-Small                 & 46.4        & 56.6         & 55.0        & 61.5         & 38.1          & 52.0           \\
T5-Base                 & 46.8        & 56.4         & 55.0        & 61.3         & 38.9          & 51.7           \\
Ours       & \bf{52.5}        & \bf{62.5}         & \bf{61.6}        & \bf{67.8}         & \bf{43.7}          & \bf{57.4}           \\ 

Ours (DPM-Solver++)                & 48.6        & 57.1         & 57.3        & 62.9         & 40.0          & 51.5          \\ 
\hline 
\end{tabular}
\caption{Results on the ToTTo development set. We obtained comparable with baselines metrics.}
\label{tab:results-totto-dev}
\end{table*}

\begin{table*}
\centering
\begin{tabular}{llllllllll} \hline
\multirow{2}{*}{Model} & \multicolumn{3}{c}{Overall} & \multicolumn{3}{c}{Overlap}  & \multicolumn{3}{c}{Non-Overlap} \\ 
& BLEU$\uparrow$ & PARENT$\uparrow$ & BLEURT$\uparrow$ & BLEU$\uparrow$  & PARENT$\uparrow$ & BLEURT$\uparrow$ & BLEU$\uparrow$ & PARENT$\uparrow$ & BLEURT$\uparrow$     \\ \hline
T5-3B \cite{kale-rastogi-2020-text} & \bf{49.5} & 58.4 & 23.0 & \bf{57.5} & 62.6 & 35.1 & \bf{41.4} & 54.2 & 10.8 \\ 
T5-base-CONT \cite{an2023cont} & 49.1 & \bf{58.9} & \bf{23.8} & 56.7 & \bf{63.2} & \bf{35.5} & 41.3 & \bf{54.6} & \bf{12.1} \\
T5-base-Lattice \cite{wang2022robust} & 48.4 & 58.1 & 22.2 & 56.1 & 62.4 & 34.5 & 40.4 & 53.9 & 9.9 \\
Pointer-Generator \cite{see-etal-2017-get} & 41.6 & 51.6 & 7.6 & 50.6 & 58.0 & 24.4 & 32.2 & 45.2 & -9.2  \\
Content Planner \cite{puduppully2019datatotext} & 19.2 & 29.2 & -57.6   & 24.5    & 32.5 & -49.1 & 13.9   & 25.8  & -66.2   \\
BERT-to-BERT \cite{Rothe_2020} & 44.0 & 52.6 & 12.1 & 52.7 & 58.4 & 25.9 & 35.1 & 46.8 & -1.7 \\
Ours           & 47.1 & 56.9 & 19.1 & 55.4 & 61.9 & 33.3 & 38.5 & 51.9 & 4.8 \\ 
\hline 
\end{tabular}
\caption{Results on the ToTTo test set. Our diffusion model outperforms BERT-to-BERT \cite{Rothe_2020} which is twice number of parameters, and achieves comparable with state-of-the-art baselines metrics. BLEURT is original BLEURT score multiplied by 100 for comparable looks to other measures.}
\label{tab:results-totto-test}
\end{table*}

\begin{table}
\centering
\begin{tabular}{@{}lcccc@{}}
\toprule
& Self-BLEU$\downarrow$ & div-4$\uparrow$  & dist-1$\uparrow$ &  Length \\ 
\midrule
Bert-to-Bert (beam) \cite{ramamurthy2023reinforcement} & 75.0 & - & - & - \\
Bert-to-Bert (top-k) \cite{ramamurthy2023reinforcement} & 82.2 & - & - & - \\
Bert-to-Bert (top-p) \cite{ramamurthy2023reinforcement} & 84.3 & - & - & - \\
T5-3B (beam) \cite{ramamurthy2023reinforcement} & 83.7 & - & - & - \\
T5-3B (top-k) \cite{ramamurthy2023reinforcement} & 88.9 & - & - & - \\
T5-3B (top-p) \cite{ramamurthy2023reinforcement} & 90.1 & - & - & - \\
PlanGen \cite{ramamurthy2023reinforcement} & \bf{25.9} & - & - & - \\
T5-small & 31.9 & 69.5 & 94.9 & \bf{87.6} \\
T5-base & 31.1 & \bf{70.7} & \bf{95.5} & 84.4 \\
Ours & 35.4 & 69.1 & 93.0 & 81.4 \\
\bottomrule
\end{tabular}
\caption{Diversity metrics on the ToTTo development set. Our model maintains diversity near the auto-regressive baselines yet. }
\label{tab:diversity}
\end{table}

\subsection{Experiment setup}

\subsubsection{Dataset}
We evaluate our method from the quality and diversity aspects on the ToTTo dataset. The validation and testing sets are based on examples from the training samples and other new tables. Test set $\mathbf{D}_{\textrm{test}}$ is divided into overlap and non-overlap sets in the following way:
\begin{equation*}
\mathbf{D}_{\textrm{test-overlap}} := \{ d : h(d) \text{ in } h(\mathbf{D}_{\textrm{train}}) \} \text{,}
\end{equation*}
\begin{equation*}
\mathbf{D}_{\textrm{test-nonoverlap}} := \{ d : h(d) \text{ not in } h(\mathbf{D}_{\textrm{train}}) \} 
\end{equation*}
where $h$ is the header value of table $d$. Development  set $\mathbf{D}_{\textrm{dev}}$ is divided into parts similarly \cite{parikh2020totto}.

\subsubsection{Baselines}
As baselines, we use T5 models with 60M (small version) and 220M (base version) parameters. We finetune them 10 epochs (about 150000 steps) with AdamW \cite{loshchilov2019decoupled} optimizer with learning rate 2e-4, weight decay 0, and batch size 8 on V100, following the paper \cite{wang2022robust}. All other baseline results on ToTTo can be found in the official leaderboard.

\subsubsection{Evaluation}
For quality evaluation, we use the metrics BLEU \cite{parikh2020totto, papineni-etal-2002-bleu} to measure intersection with the target sequence by n-grams, PARENT \cite{parikh2020totto, dhingra-etal-2019-handling} to get the F1-score with table data, and BLEURT \cite{sellam2020bleurtlearningrobustmetrics} to robust evaluate closeness of the target sequence and the predictions on overall, overlapping and non-overlapping subsets. To assess diversity, we generate 10 samples and use three metrics such as div-4 \cite{gong2023diffuseq, deshpande2019fast} to measure the percentage of distinct 4-grams in the set of outputs, \text{dist-1} \cite{gong2023diffuseq} to measure internal diversity as the percentage of distinct unigrams, and self-BLEU \cite{gong2023diffuseq, zhu2018texygen} as the metric of overlapping between different candidates. 

Also, we compute the average length of predictions to understand the reasons for changing BLEU and diversity metrics. In all tables, the length is presented in the number of characters.

\subsubsection{Implementation Details}
As a core model we use the official implementation GENIE \cite{lin2023text} \footnote{\url{https://github.com/microsoft/ProphetNet/tree/master/GENIE}}. Our model is a \text{6-layer} transformer as the encoder and a 6-layer cross-attention transformer as the denoising decoder. The latent variable dimension to 768 and the embedding dimension to 128. Our model has 144M parameters. The batch size is 64. The input length is 475 and the output length is 119. The model was trained for 120000 steps with the help of AdamW optimizer \cite{loshchilov2019decoupled} with a learning rate 5e-5 and weight decay 0 on V100. After the training, we compute the exponential moving average of the model's weights for stability.

\subsection{Main results}
We compare the iter-NAR encoder-decoder diffusion model with modern text-to-text baselines T5-small and T5-base \cite{kale-rastogi-2020-text}. For the quality of generation, we have followed the existing work \cite{parikh2020totto}, reporting BLEU \cite{papineni-etal-2002-bleu} and PARENT \cite{dhingra-etal-2019-handling} on the Overall, Overlap, and Non-Overlap parts of the development and test parts of the dataset ToTTo \cite{parikh2020totto}. For the test part we also public learning robust metric BLEURT \cite{sellam2020bleurtlearningrobustmetrics} for completeness of comparison. We evaluated the diversity of these models following the \cite{gong2023diffuseq} paper by length, dist-1, div-4 \cite{deshpande2019fast}, and Self-BLEU \cite{zhu2018texygen} metrics.

We present the main results of the diffusion model and the baselines on the dataset ToTTo in Fig. \ref{tab:results-totto-dev} and Fig. \ref{tab:diversity} for the dev set, and in Fig. \ref{tab:results-totto-test} for the test set. The results are submitted to the ToTTo leaderboard \footnote{\url{https://github.com/google-research-datasets/ToTTo}} - \text{Diffusion-TT}.

\subsubsection{Quality}
For the development set, to evaluate the quality of the diffusion model, we sample 10 examples from our model and choose the best prediction. For T5 models decoding is done via the usual greedy search following the paper \cite{kale-rastogi-2020-text}. For the test set, we apply Minimum Bayes-Risk (MBR) \cite{gong2023diffuseq, inproceedings} on 30 predictions (10 of each model with different output lengths - 119, 128, 153) to achieve the best results. Our results demonstrate that the diffusion model can achieve comparable quality with T5 models. Moreover, we see that for all models, the difference in metrics between the Overlap and Non-Overlap parts of the dataset is essentially the same.  From this, we might infer that overfitting affects all models to a roughly equal degree.

\subsubsection{Diversity}
From the table \ref{tab:diversity} can see that the diffusion model achieves comparable diversity metrics, but falls behind the plan-based model \cite{ramamurthy2023reinforcement} and auto-regressive models with temperature sampling. We see that our model confidently outperforms auto-regressive models with other types of diverse generations such as beam search \cite{xiao2023survey}, nucleus \cite{holtzman2020curious}, and top-k generation \cite{fan2018hierarchical}. Also, we note that auto-regressive models generate longer predictions than our diffusion model, so it may be the cause for the slightly higher diversity metrics. The examples of both models generation are presented in Fig. \ref{tb:case}.

\subsection{Ablation study}

\begin{table*}
\centering
\begin{tabular}{lllllll} \hline
\multirow{2}{*}{Model} & \multicolumn{2}{c}{Overall} & \multicolumn{2}{c}{Overlap}  & \multicolumn{2}{c}{Non-Overlap} \\ 
                      & BLEU$\uparrow$        & PARENT$\uparrow$        & BLEU$\uparrow$        & PARENT$\uparrow$        & BLEU$\uparrow$          & PARENT$\uparrow$          \\ \hline
T5-Small                 & 36.8        & 48.3         & 44.8        & 53.7         & 29.1          & 43.1           \\
T5-base                 & 37.5        & 47.8         & 45.2        & 53.2         & 30.1          & 42.5          \\
Ours                  & \bf{38.6}        & \bf{49.0}         & \bf{46.5}        & \bf{54.5}         & \bf{30.7}          & \bf{43.7}           \\ 

Ours (DPM-Solver++)                & 41.5        & 50.6        & 49.7        & 56.6         & 33.4         & 44.9          \\ 
\hline 
\end{tabular}
\caption{Impact of different samplers on the ToTTo development set. All metrics are estimated on 1 sample.}
\label{tab:fastsampler_quality}
\end{table*}

\Figure[h](topskip=0pt, botskip=0pt, midskip=0pt){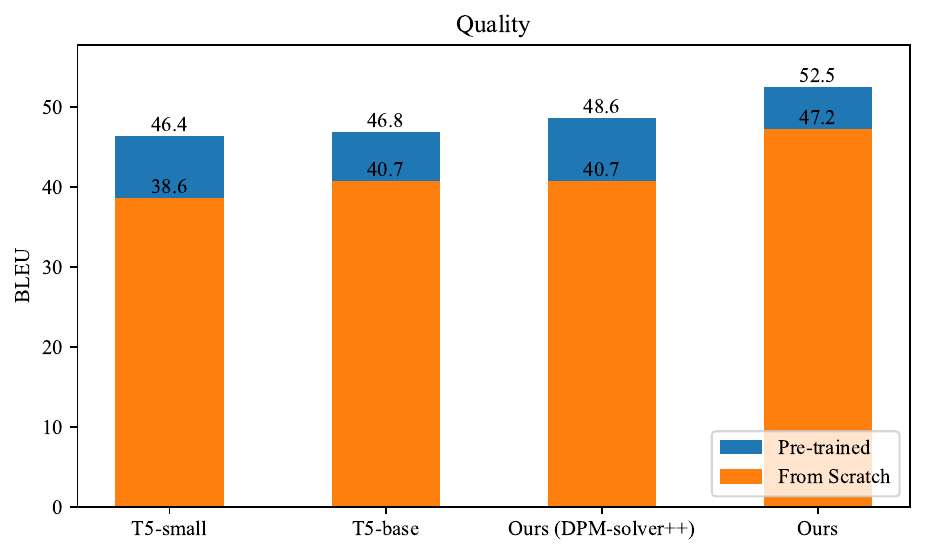}
{ \textbf{Difference between pre-trained and trained from scratch models. The result of our model is the optimal
value of 10 Gaussian samples.}\label{fig:scratch}}

\subsubsection{Impact of different samplers}
DPM-Solver++ \footnote{\url{https://github.com/Shark-NLP/DiffuSeq/tree/diffuseq-v2}} \cite{gong2023diffuseqv2} was adopted for text-to-text diffusion models and increased the speed of training by 4 times and the speed of inference by 800 times without loss of quality \cite{gong2023diffuseqv2}. We applied it to our encoder-decoder diffusion model. We see that this solver increases quality metrics, expressed in Fig. \ref{tab:fastsampler_quality}, but, diversity metrics and the length have deteriorated greatly (Fig. \ref{tab:fastsampler_diversity}). The model starts generating nearly identical, brief, and highly accurate sentences, which when aggregated, do not improve model quality.
\begin{table}
\centering
\begin{tabular}{@{}lcccc@{}}
\toprule
& Self-BLEU & div-4  & dist-1 &  Length \\ 
\midrule
Ours & 35.4 & 69.1 & 93.0 & 81.4 \\
Ours with Fast Sampler & 69.6 & 30.3 & 94.7 & 71.1\\
\bottomrule
\end{tabular}
\caption{Impact of different samplers. Results on the ToTTo development set. DPM-Solver++ \cite{gong2023diffuseqv2} reduces diversity, but improves the quality of each prediction.}
\label{tab:fastsampler_diversity}
\end{table}

\subsubsection{Prediction aggregation methods}
The challenge with diffusion models for text is that we consistently receive various predictions and struggle to obtain the optimal one every time.
We can use different strategies to aggregate predictions in text diffusion models. We test ROVER \cite{659110}, MBR \cite{gong2023diffuseq, inproceedings}, and Best-of-all \cite{lin2023text} algorithms on our model and T5. All quality metrics correlate in this experiment, we only use BLEU in Fig. \ref{tab:aggreg}. We see that the best results without the target's leak are obtained with the help of the MBR.\cite{inproceedings}.

\begin{table}
\centering
\begin{tabular}{@{}llccc@{}}
\toprule
&  & MBR  & ROVER  & Best-of-all  \\
\midrule
 & T5-base & \bf{45.5} & 39.3 & \bf{54.3} \\
 & T5-small & 45.4 & 39.7 & 53.9 \\
 & Ours & 44.6 & 40.7 & 52.5 \\
 & Ours (DPM-Solver++) & 42.7 & \bf{41.3} & 48.6 \\
\bottomrule
\end{tabular}
\caption{Comparison of different prediction aggregation methods. All algorithms are applied to 10 samples.}
\label{tab:aggreg}
\end{table}

\subsubsection{Impact of sampling temperature}
We also study the impact of sampling temperature in T5 models on diversity in comparison to diffusion models.
The decrease of the temperature in the sampling of T5 leads to the quality increase but diversity deterioration. In Fig. \ref{tab:temp} we explore how diversity and quality depend on sampling temperature in T5 vs our model. The key observation is that our diffusion model demonstrates the balance between quality and diversity whilst T5-models with the temperature near one get worse in quality and near zero get worse in diversity.

\begin{table*}
\centering
\begin{tabular}{@{}llccc@{}}
\toprule
&  & Self-BLEU$\downarrow$ & BLEU-Overall$\uparrow$ & PARENT-Overall$\uparrow$ \\
\midrule
 & T5-small-1 & 31.9 & 36.8 & 48.3 \\
 & T5-small-0.97 & 33.3 & 37.6 & 49.0 \\
 & T5-small-0.85 & 40.7 & 41.0 & 51.9 \\
 & T5-small-0 (greedy) & 100.0 & 46.4 & \bf{56.6} \\
 & T5-base-1 & \bf{31.1} & 37.5 & 47.8 \\
 & T5-base-0.97 & 32.3 & 38.1 & 48.1 \\
 & T5-base-0.85 & 37.5 & 40.7 & 50.6 \\
 & T5-base-0 (greedy) & 100.0 & \bf{46.8} & 56.4 \\
 & Ours & 35.4 & 38.6 & 49.0 \\
 & Ours (DPM-Solver++) & 69.6 & 41.5 & 50.6\\
\bottomrule
\end{tabular}
\caption{Impact of sampling temperature to BLEU \cite{papineni-etal-2002-bleu} on the overall development set and self-BLEU \cite{zhu2018texygen} between 10 samples.}
\label{tab:temp}
\end{table*}

\subsubsection{Pre-training impact}

The pre-training is known to increase generalization in text-to-text tasks. We evaluate if it matters on our task. We compared auto-regressive models vs diffusion models with and without pre-training.

As we see in Fig. \ref{fig:scratch} and in Fig. \ref{tab:scratch-quality}, diffusion models trained from scratch outperform auto-regressive baselines. We believe it is caused by increasing size of the training dataset with help of the adding gaussian noise.

However, the pre-training improves quality for all models evenly, resulting in a 15 to 16 percent increase.

\begin{table*}
\centering
\begin{tabular}{lllllll} \hline
\multirow{2}{*}{Model} & \multicolumn{2}{c}{Overall} & \multicolumn{2}{c}{Overlap}  & \multicolumn{2}{c}{Non-Overlap} \\ 
                      & BLEU$\uparrow$        & PARENT$\uparrow$        & BLEU$\uparrow$        & PARENT$\uparrow$        & BLEU$\uparrow$          & PARENT$\uparrow$          \\ \hline
T5-Small                 & 38.6        & 47.3         & 47.2        & 53.2         & 30.1          & 41.6           \\
T5-Base                 & 40.7        & 51.3         & 50.5        & 57.6         & 31.5          & 45.1           \\
Ours                  & \bf{47.2}        & \bf{57.5}         & \bf{56.5}        & \bf{63.6}         & \bf{38.1}          & \bf{51.5}           \\ 

Ours (DPM-Solver++)                & 40.7        & 51.0         & 48.8        & 56.9         & 32.6          & 45.4          \\ 
\hline 
\end{tabular}
\caption{Results on the ToTTo development set for trained from scratch models. Diffusion models outperform auto-regressive baselines in quality metrics.}
\label{tab:scratch-quality}
\end{table*}

\subsubsection{Impact of output length}

We explore how the fixed length of outputs influences quality and diversity.  We select three various lengths, 119 as the maximum length of the target sequence in the training subset ToTTo \cite{parikh2020totto}, 128 as the classical output length of our baselines, and 153 as the length in the pre-training phase of our diffusion model \cite{lin2023text}. Our diffusion model generates better results with the small length, as we see in Fig. \ref{tab:length} and Fig. \ref{fig:mesh3}. 
We also observe that the average output prediction length did not change with the increasing maximum allowed output length.
The diffusion model uses normal noise during the generation process, so the longer the length, the more noise the model receives and the higher the likelihood of receiving an inappropriate word. This, we believe, explains why the quality of the model predictions decreases with length. This leads to an increase in inaccuracy, but it appears that these hallucinations only alter words, they do not add new ones, because the average length remains the same.
\Figure[t!](topskip=0pt, botskip=0pt, midskip=0pt){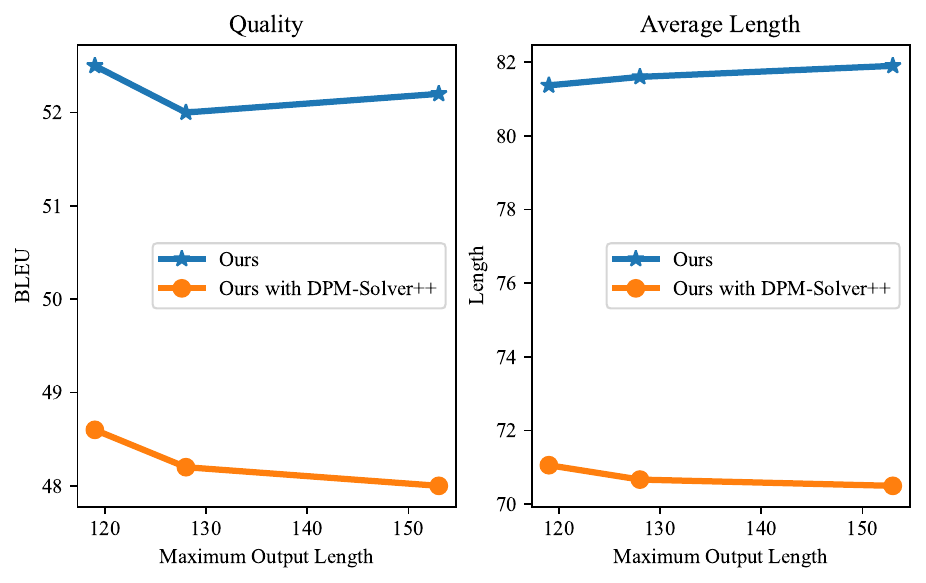}
{ \textbf{Effect of the maximum output length on the ToTTo development set. The result is the optimal
value of 10 Gaussian samples.}\label{fig:mesh3}}

\begin{table}
\centering
\begin{tabular}{@{}lccc@{}}
\toprule
& Maximum Output length  & BLEU $\uparrow$ & Length  \\
\midrule
 Ours  & 119 & \bf{52.5} & 81.4 \\
 & 128 & 52.0 & 81.6 \\
     & 153 & 52.2 & 81.5 \\
Ours (DPM-Solver++) & 119 & 48.6 & 71.1 \\
     & 128 & 48.2 & 70.7 \\
     & 153 & 48.0 & 70.5 \\
\bottomrule
\end{tabular}
\caption{Effect of the maximum output length on the ToTTo development set. The result is the optimal
value of 10 Gaussian samples.}
\label{tab:length}
\end{table}

\section{Conclusion and future work}

We investigated diffusion models application to the table-to-text problem by adapting them to the task and conducting an in-depth analysis. Our experiments covered multiple aspects of diffusion models training. We explored sampling strategy influence by inducing recent diffusion model accelerator DPM-Solver++ \cite{gong2023diffuseqv2} into our core model and found out that DPM-Solver++ \cite{gong2023diffuseqv2} reduces diversity, but improves the quality of each prediction. In addition, we assessed many prediction aggregation techniques, including Minimum Bayes Risk (MBR) and ROVER, and found that MBR was more effective. We also looked into the effects of generation length limits and the pre-training phase on diffusion models. It was discovered that diffusion models trained from scratch outperformed auto-regressive baselines. In order to assess variety, we also contrasted the performance of diffusion models with that of auto-regressive text-to-text models at various temperature settings. The key observation is that our diffusion model demonstrates the balance between quality and diversity whilst T5-models with the temperature near one get worse in quality and near zero get worse in diversity. Our findings reveal that diffusion models achieve comparable results in the table-to-text domain, highlighting their diversity in the table-to-text challenge as a promising research direction. 
We see the future of this topic to explore more modern transformers variations at the core of the model along with the scaling experiments. Another path of investigation is the effect of resource allocation, the impact of the text structure and table complexity.

\section{Limitations}
We use the pre-trained model GENIE \cite{lin2023text}, which has the vanilla BERT-to-BERT \cite{Rothe_2020} transformer core. We did not test other contemporary architectures like T5 \cite{raffel2023exploring} or Deberta-V3 \cite{he2023debertav3}.  




\bibliography{main}
\bibliographystyle{splncs04}

\begin{IEEEbiography}[{\includegraphics[width=1in,height=1.25in,clip,keepaspectratio]{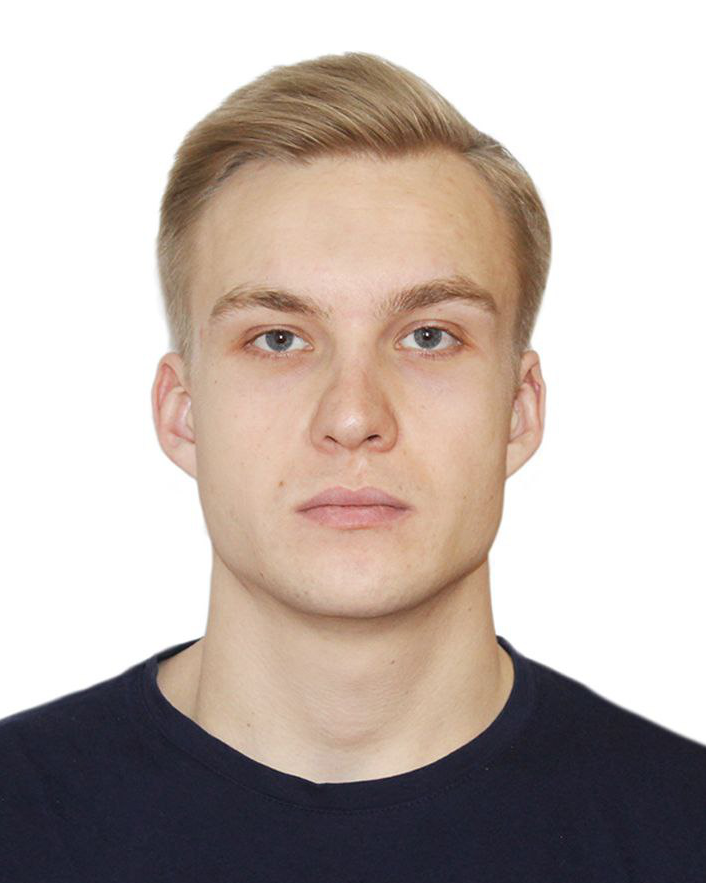}}]{Aleksei S. Krylov} received the B.S. degree in applied mathematics and computer science from Moscow Institute of Physics and Technology, Moscow, Russian Federation, in 2022 and M.S degree in
machine learning and data analysis from Moscow Institute of Physics and Technology, Moscow, Russian Federation, in 2024.
In 2020, he was a Summer Intern at Intel. From 2020 to 2022, he was a Researcher with the Laboratory of Information Technologies and Applied Mathematics at Moscow Institute of Physics and Technology. Since 2022, he has been a data scientist in the biggest bank in Russian Federation, Sber.
His research interests include natural language processing and machine learning in risk management.
\end{IEEEbiography}

\begin{IEEEbiography}[{\includegraphics[width=1in,height=1.25in,clip,keepaspectratio]{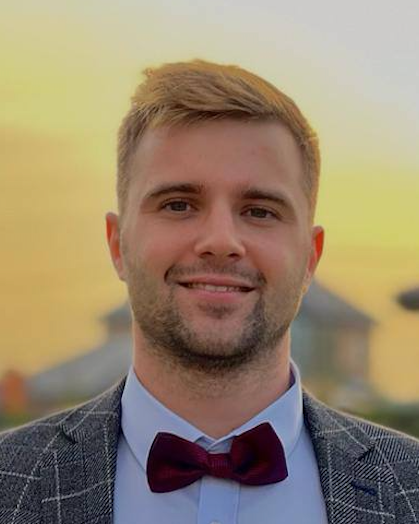}}]{Oleg D. Somov} received the B.S. degree in applied mathematics and computer science from Moscow Institute of Physics and Technology, Moscow, Russian Federation, in 2018 and M.S degree in
machine learning and data analysis from Moscow Institute of Physics and Technology, Moscow, Russian Federation, in 2020.
Currently PhD in progress in Moscow Institute of Physics and Technology, Moscow, Russian Federation. Currently Oleg works as a research scientist at Artificial Intelligence Research Institute (AIRI), Moscow, Russian Federation.
\end{IEEEbiography}

\EOD

\end{document}